# Design of an Affordable Prosthetic Arm Equipped with Deep Learning Vision-Based Manipulation

Alishba Imran[1], William Escobar[1], Dr. Freidoon Barez[1]

[1]San Jose State University, San Jose, CA

*Abstract*—Many amputees throughout the world are left with limited options to personally own a prosthetic arm due to the expensive cost, mechanical system complexity, and lack of availability. The three main control methods of prosthetic hands are: (1) body-powered control, (2) extrinsic mechanical control, and (3) myoelectric control. These methods can perform well under a controlled situation but will often break down in clinical and everyday use due to poor robustness, weak adaptability, long-term training, and heavy mental burden during use.

This paper lays the complete outline of the design process of an affordable and easily accessible novel prosthetic arm that reduces the cost of prosthetics from $10,000 to $700 on average. The 3D printed prosthetic arm is equipped with a depth camera and closed-loop off-policy deep learning algorithm to help form grasps to the object in view. Current work in reinforcement learning masters only individual skills and is heavily focused on parallel jaw grippers for in-hand manipulation. In order to create generalization which better performs real-world manipulation, the focus is specifically on using the general framework of Markov Decision Process (MDP) through scalable learning with off-policy algorithms such as deep deterministic policy gradient (DDPG) and to study this question in the context of grasping a prosthetic arm. We were able to achieve a 78% grasp success rate on previously unseen objects and generalize across multiple objects for manipulation tasks.

## I. Introduction

There are an estimated 30 million people needing prosthetic devices with a shortage of 40,000 prosthetic devices in developing countries. Prosthetic arms today are expensive, with costs of $5,000 for a purely cosmetic arm, up to $10,000 for a functional mechanical prosthetic arm that ends in a split hook, and up to $20,000 to $100,000 or more for an advanced myoelectric arm controlled by muscle movements. High cost, poor design, and technology has also led to a shortage in trainees for training which can often take 3 or more months [1].

### A. Body-Powered Mechanical Control

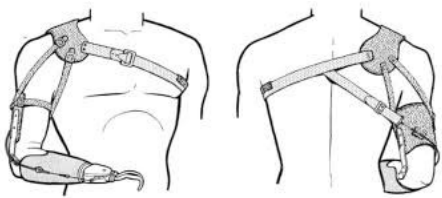

Figure 1. Mechanically Actuated Prosthetic Arm [2]

Body-powered prostheses are currently the most used functional artificial arm due to their simplicity. A body-powered prosthetic arm is held in place by suction or by a strap system that utilizes your shoulders. A system with a cable and harness are the main way of controlling the device [3].

There are two main types of body-powered hand prostheses:
- Voluntary Open: opens the hand when applying tension to the cable.
- Voluntary Close: closes the hand when applying tension to the cable.

Barriers/Current Problems:
- Cost: ~$30,000
- Rejection rate 16–58% (often very uncomfortable)
- Restrict the range of movement of arm.
- Difficult to finely control grasp and grip.

### B. Externally Myoelectric Sensors

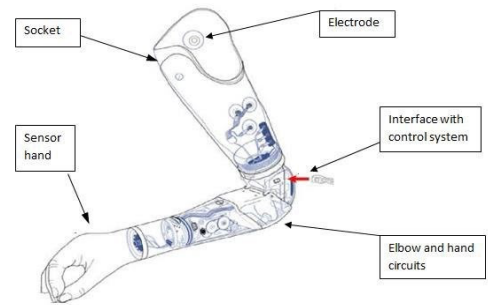

Figure 2. Myoelectric Prosthetic Arm [2]

Myoelectric prostheses do not require a harness and are controlled by nerves and electrical impulses from the wearer's residual limb. When muscles are moved, small electrical fields are generated around them, which can be measured by electrodes. Sensors or electrodes in the prosthetic socket detect your muscle contractions and send commands to operate the high performance, battery-operated prosthetic motors. They often use two electrode sites to sense muscle contractions from two major muscle groups.

Barriers/Current Problems:

- Cost: ~$100,000 (expensive)
- Batteries need to be recharged often.
- Lengthy training period to adapt.
- Difficult to finely control grasp and grip.

*C. Implanted Myoelectric Sensors (IMES)*

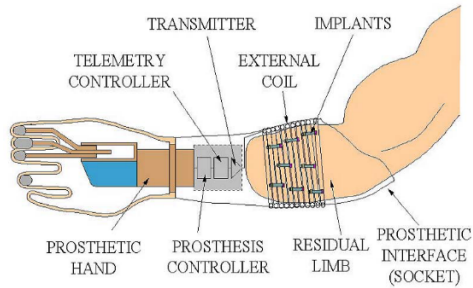

Figure 3. Implanted Myoelectric Prosthetic Arm [1]

The IMES system is a device used to improve signal quality and consistency of myoelectric signals for prosthetic control. The end caps serve as electrodes for picking up EMG activity during muscle contraction. Reverse telemetry (via a coil around the arm) is used to transfer data from the implanted sensor, and forward telemetry is used to transmit power and configuration settings to the sensors. The coil and associated electronics are housed within the frame of a prosthesis. A control system that sends data associated with muscle contraction to the motors of the prosthetic joints is housed in a belt-worn, battery-powered device. A cable attaches the control unit to the prosthetic frame. An IMES is implanted into each targeted muscle that will be used to control a function of the prosthetic arm. This means, for each function would need one implanted sensor in one muscle per function (open hand or close hand) [4].

Barriers/Current Problems:
- Total Cost: ~$1,382,120 (very expensive)
- Invasive Process

The expensive cost and accessibility of the current prosthetic arms leaves many amputees without the option to personally own a prosthetic arm throughout the world.

## II. RELATED WORK

*A. Features and Specifications of Prosthetic Arms*

An empirical study [5] of the many features and specifications of many different types of prosthetic arms has been performed. The report examined finger design and kinematics, mechanical joint coupling, and actuation methods. The arms that were evaluated were Vincent, iLimb, iLimb Pulse, Bebionic, Bebionic v2, and Michelangelo prosthetic hands.

After the study, the article determined that the most desirable features of a prosthetic arm are:

- Total weight < 500 g
- Simple and robust designs are better than anatomically correct designs.
- Thumb Powered adduction to achieve different grips.
- Use of brushless motors over brushed motors for performance.
- Pinch force at fingertips to be maxed out at 65 N in palmar prehension.
- Minimal acceptable angular speed should be at 115 degrees/sec.
- All functional grasping hands should be included in the design with a low number of actuators with transmissions.

In this study, these benchmark features were used as starting point to create the first initial prototype.

*B. 3D Printed Prosthetics*

3D printing is becoming an integral part of upper-limb prosthesis as a way to increase access to prostheses in a timely manner. The ability to rapidly produce models has been shown to be effective in the clinical setting [6] but 3D-printed parts often raise concerns of safety in daily use [7].

Many organizations such as e-NABLE, Robohand, and Limbitless Solutions have created several open source models of body-powered 3D printed prosthetic hands. While the e-NABLE hand can perform multiple tasks, it has difficulty accomplishing others as the prosthetic hand is designed so that all the fingers bend together causing every finger to move in line with the rest. This makes it difficult for every finger to be able to touch the object. Furthermore, a lot of 3D printed prosthetics aren't able to handle large amounts of load as the material will break down on rigorous day to day use.

In this study, we focus on developing a safe prosthetic arm that is 3D printed with a modular design to handle a large load and that is able to grasp multiple everyday objects in a flexible way.

*C. Deep Learning Approaches*

Current work for grasping tasks is heavily focused on parallel jaw grippers for in-hand manipulation of robots [8], [9]. Although parallel jaw grippers can be used for grasping, just grasping an object is not a very useful skill by itself as humans perform complex in-hand manipulation of grasped objects for achieving desired goals.

Past work on controlling anthropomorphic hands has relied on imitation [10], human demonstration [11], [12] or pre-defined grasp types [10] for control strategies. As the observation space becomes richer and the complexity of tasks increases, using human demonstrations for learning control policies is not scalable because collecting demonstrations is known to be a hard and tedious process.

Reinforcement learning (RL) offers a promising avenue for tackling this problem, but current work on reinforcement learning masters only individual skills. With RL, it is more difficult to find a policy to control a prosthetic arm in a desirable manner as opposed to parallel jaw grippers for in-hand manipulation. To create generalization which better meets of real-world manipulation, we focus specifically on using the general framework of Markov Decision Process (MDP) through scalable learning with off-policy algorithms such as Deep Deterministic Policy Gradient (DDPG) [13] and study this question in the context of the specific problem of grasping a prosthetic arm.

## III. MATERIALS AND METHODS

### A. Modeling Prosthetic Arm with Solidworks

For a prosthetic arm, the size specifications should be similar in size of the user, therefore the first prototype was made by measuring the lengths and width of one of our hands. Table 1. shows all of the dimensions taken with calipers.

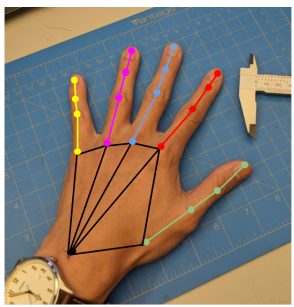

Figure 4. Hand Measurements

TABLE I. FINGER AND HAND DIMENSIONS

| Dimensions mm | Wrist to 1st Joint (Palm) | 1st Link | 2nd Linkage | 3rd Linkage |
|---|---|---|---|---|
| Thumb | 52 | 45 | 34 | 29.5 |
| Index | 90 | 43.5 | 26.5 | 23.5 |
| Middle | 86 | 46.5 | 32 | 24.5 |
| Ring | 79 | 45.5 | 32 | 24 |
| Pinky | 79 | 35.5 | 23.5 | 21 |

### B. Finger Assembly Mechanism Minimum Servo Torque Calculations for Grip Force

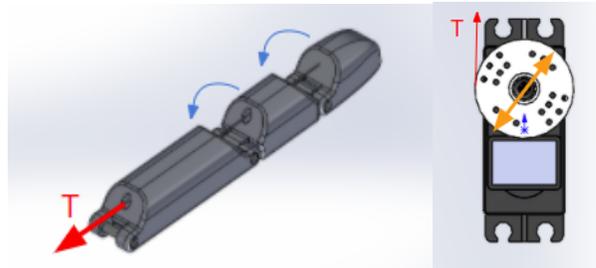

Figure 5. Typical Tendon Mechanism with Servo to Apply Tension

The pinch force of 65N, Fg, from the benchmark specifications and free body diagram can be used to calculate an estimate minimum torque requirement needed from the servos.

The knuckle joint to the tip of the fingertip, R, is a function of servo angle position. We can calculate the minimum servo torque at the near close position for pinch force. Assumption of the assembly is that it is frictionless.

Consider the following free body diagram in pinch state:

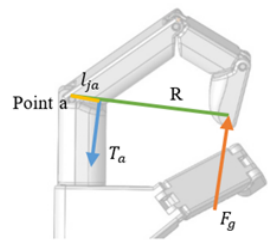

Figure 6. Free Body Diagram of Finger Assembly

Taking the sum of moments about point a:

$$\Sigma M_a = 0 \quad (1)$$

Breaking it further into components:

$$(F_g)(R) - (T_a)(l_{ja}) = 0 \quad (2)$$

Where: $F_g = 65N$; $R = 53mm$; $l_{ja} = 5mm$

Solving for $T_a$:

$$T_a = 689 \; Nmm \quad (3)$$

Minimum servo torque can be calculated:

$$\tau = \frac{d}{2} T_b \sin\theta \quad (4)$$

Where: $T_b = -T_a = -689 \; Nmm$; $d = 23.8mm$;

$$\theta = 90°$$

$$\tau = -8199 \; Nmm \quad (5)$$

## C. Controller Design with MATLAB and Simulink

Using MATLAB's Simulink, we can help design a controller that will help us prevent from the prosthetic arm grip from overshooting the setpoint.

Future work includes constructing the transfer function of the finger assembly, the servo, and the force sensors. Once these transfer functions are completed, it will be inputted into the Simulink graphical user interface and using analytical tools, a control system will be developed.

## D. Finite Element Analysis with 3D Printed PLA Orthotropic Material

The main manufacturing process for custom parts are made with a 3D FDM printer. Although 3D printers offer quick turn around on parts, the parts will end up being transversely isotropic, or orthotropic, due to the nature of the FDM printing.

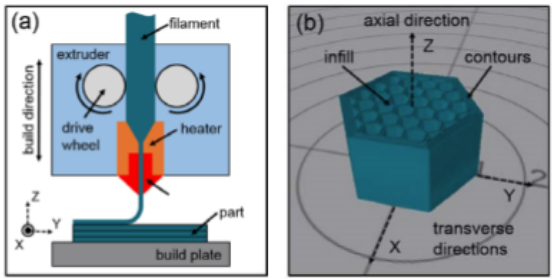

Figure 7. FDM 3D Printing Process

It is important to include the directions of the print in FEA to accurately represent the strength of material in all directions.

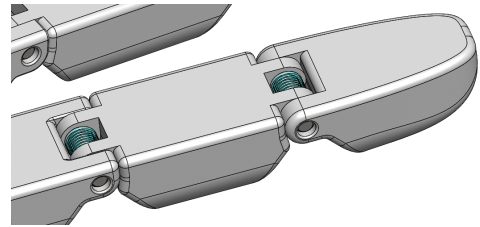

Figure 8. Transversely Orthotropic PLA Material Properties

## IV. RESULTS AND DISCUSSION

### A. First Prototype

The first 8-DOF prototype includes 6 servos (5 for each finger, 1 for thumb abduction) with the typical tendon mechanism method that runs tension through each finger (stiff string) to contract each finger to grasp the object. The decision and motivation behind this mechanism from other choices is due to the low cost. The assembly method applied for the joints of each finger assembly is press fit dowl pins. There are also torsional springs in each joint to have an open hand at the resting state when no tension is applied to each tendon.

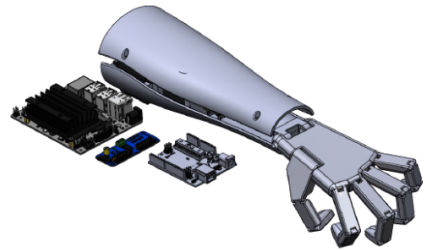

Figure 9. First CAD Model Iteration

Figure 10. Press Fit Dowel Pin and Torsional Spring

### B. FEA Results

To validate the prototype design and determine the structural integrity, Ansys was used to simulate force loads on assumed weak areas of the assembly. The force loads that are simulated are from payload and forces due to the servos.

The simulation was separated into three parts to evaluate two joints, which are the wrist joint and the knuckle joint, and the press fit process for dowel pins.

The process to fine the load acceptable for each analysis were to set a parameter of the force load where it would incrementally increase and return a safety of factor value. When the safety of factor approached downward close to 1, then the force value was noted.

TABLE II. FEA FOR DOWEL PRESS FIT MAX STRESS AND SAFETY FACTOR

| Global Direction | Max Stress | Safety Factor |
|---|---|---|
| Y-Axis | 37.8 | 3.00 |
| Z-Axis | 80.0 | 1.35 |
| X-Axis | 81.0 | 1.30 |

### C. Algorithm for Manipulation: Environment and Set Up

Reinforcement learning provides a general framework for determining control policies without having to directly model the actuator or the environment. However, most current reinforcement learning algorithms [14], [15] rely on random walk in the action space for exploration making controlling actuators with large number of degrees of freedom important. In this work, we further explore learning robust grasping behavior from scratch using model free reinforcement learning with a anthropomorphic hand in a changing environmental for various different objects and grasp types.

The current study was focused on the cylindrical grasp for mutli-object manipulation, but our goal is to apply this approach to be able to form common grasps that are used in everyday settings.

- Cylindrical Grasp
- Tip
- Hook or Snap
- Palmar
- Spherical Grasp
- Lateral Pinch

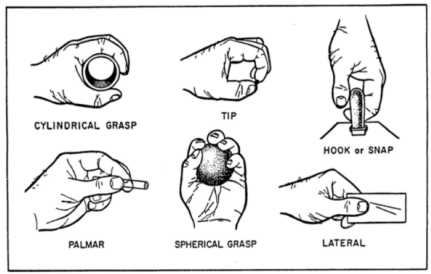

Figure 11. Common Grasps to Consider for Prosthetic Arm to Form

To complete this simulation, we use a simulated model of the anthropomorphic hand used as part of SouthHampton Hand Assessment Procedure (SHAP) [16] procedure which was established for evaluating prosthetic hands and arms. This simulated prosthetic hand could theoretically perform all useful human hand movements and was built using the Mujoco physics engine. The hand has five fingers and 25 joints out of which thirteen are actuated. Out of these thirteen, ten joints control the motion of fingers and the other three control the rotation of the hand. Additionally, there are overall 16 degrees of actuation. The hand is controlled by setting the position of these 16 actuators. The input to the model consisted of the internal position and velocity of the 25 joints and the object position resulting in 53 dimensions.

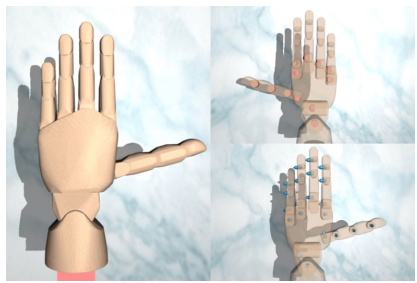

Figure 12. Design of Modular Prosthetic Limb and SHAP test suites

### D. Grasping MDP: Action, State and Reward Evaluation

Our vision-based control framework is based on a general formulation of robotic manipulation as a MDP. The policy observes the image from the robot's camera and chooses a gripper command. An MDP consists of states, actions, and a reward function. For each time step in this MDP, the accepted actions and the zoom levels of the image captured are different. Thus, we factor the Q-function into several independent CNNs, one for each time step.

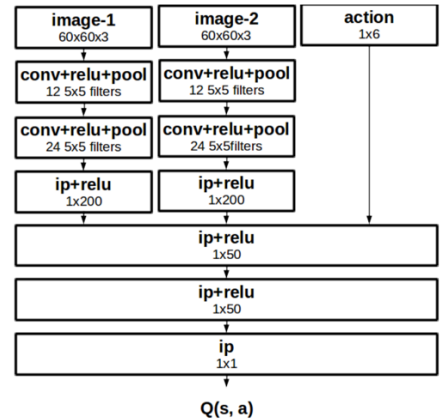

Figure 13. CNN architecture used to approximate Qt(s, a) for each step t in a pick-place process [17]

- **State**: To provide image observations, we mounted the camera on top of the prosthetic hand. We want to use images instead of point clouds to ensure that the state encoding more compact and to allow us to work with more robust convolutional neural network (CNN) architectures. For many pick-place tasks k = 2 is sufficient since there are two observations which represent (a) the target of the grasp/place and (b) what the hand contains.
- **Action**: The agent's action comprises of a gripper pose displacement which is the difference between the current pose and the desired pose in Cartesian space, and an open/close command.
- **Reward Structure**: The grasping task is defined simply by providing a reward to the learner during data collection: a successful grasp results in a reward of 1, and a failed grasp a reward of 0. We consider a grasp successful if the robot holds an object above a certain height at the end of the episode. In order to detect a successful grasp, an image subtraction test is used. Two images are captured: (a) An image of the scene after the grasp attempt, (b) A second image is taken after attempting to drop the grasped object. If no object was grasped, these two images would be identical. Otherwise, if an object was picked up, certain pixels in the two images will be different.

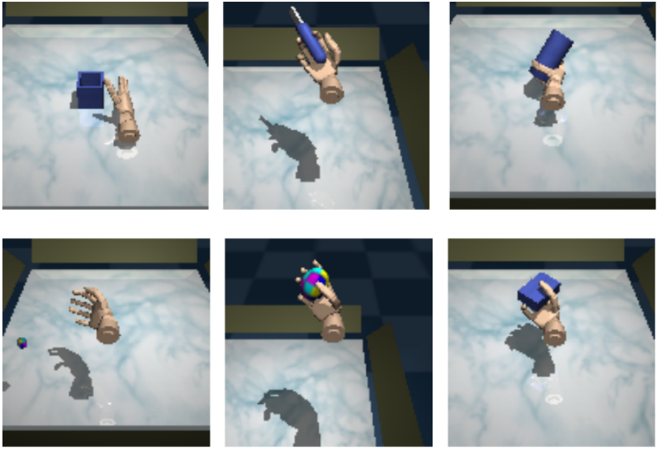

Figure 14. Some of the objects and their grasps by the model

### E. DDPG: Learning Algorithm

The framework of MDPs provides a general and powerful formalism for such decision making problems, but learning in this framework can be challenging. We trained the off-policy method of deep deterministic policy gradient (DDPG) for learning of this task.

DDPG combines elements of Q-learning [18] and policy gradients. It aims to learn a deterministic policy $\pi_\emptyset(s) = a$ by propagating gradients through a critic $Q_\theta(s, a)$. The actor is trained in DDPG according to the objective which is an approximate maximizer of the Q-function with respect to the action at any given state $s$. The practical implementation of DDPG closely follows that of Q-learning, with the addition of the actor update step after each Q-function update.

### F. RGB-D Dataset

To train our network, we built a dataset on top of the Cornell Grasping Dataset [18]. The Cornell Grasping Dataset contains 885 RGB-D images of real objects, with labelled positive and negative grasps. In order to enable our model to learn generalizable strategies that can pick up new objects, perform pre-grasp manipulation, and handle dynamic disturbances with vision-based feedback, we must train it on a sufficiently large and diverse set of objects. This dataset is effective as multiple labelled grasps are provided per image for various different objects. We augment the Cornell Grasping Dataset with random crops, zooms and rotations to create a greater set of over 140k depth images.

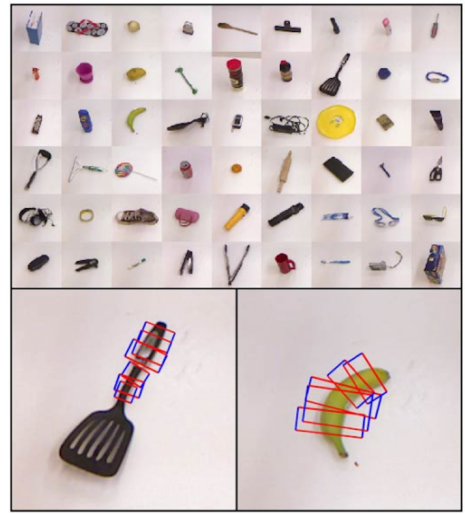

Figure 15. The Cornell Grasping Dataset [18] contains a variety of objects each with multiple labelled grasps

### G. Multiple Objects and Generalization

Once we found that it was possible to grasp a single object (cylinder), we tried to generalize the grasps to previously unseen objects. We then setup an experiment with multiple objects such as a sphere, screw driver, a can, an ellipsoid and a cylinder. We gave the learnt model novel objects that was able to learn a policy to grasp for multiple objects as well.

### H. Data efficiency

To test the robustness of training to batch size, we used batch sizes of - 60k, 80k, 120k and 140k. As batch size was increased the model learnt a more robust policy for grasping and lifting the object.

TABLE III. OBJECT GRASP SUCCESS RATE

| Object Type | Success Rate |
|---|---|
| cuboid | 0.94 |
| sphere | 0.87 |
| ellipsoid | 0.71 |
| cylinder | 0.65 |
| can | 0.76 |
| coin | 0.89 |
| screwdriver | 0.62 |

We found that this approach was effective in generalizing across multiple everyday objects such as a cuboid, sphere, ellipsoid, cylinder, can, and screwdriver. Although we did find that objects with geometric properties not previously seen had a lower success rate than the ones that were seen and trained on before. We were able to achieve a 78% grasp success rate on previously unseen objects and 84% on household objects that are moved during the grasp attempt.

## V. NEXT STEPS

We're working on setting up a simulation to test this approach on our custom built prosthetic arm design (Figure 9)

in ROS and then further test it on the 3D printed prosthetic arm in a lab setting. We want to further benchmark how our approach performs compared to existing RL approaches such as Q-learning [19], Monte Carlo [20], and trust region policy optimization (TRPO) [21]. In continuing conversations with rehabilitation professionals, we want to get more feedback and testing done on amputees to improve the overall design and use-cases for the prosthetics arms to support amputees on day-to-day activities.

## VI. Conclusion

This paper studied the design of a cheaper 3D printed prosthetic arm using a depth camera and closed-loop off-policy deep learning algorithm to help form grasps. In order to keep the cost to a minimum, a 3D printer that utilizes the method of fused deposition modelling (FDM), is used to manufacture custom parts.

Furthermore, servo actuators and force sensors are used and limited to off the shelf parts to maintain high accessibility. MATLAB and Simulink are used to help with the development of the hand control systems by modelling the equations of motion and developing a controller in order to prevent gripping an object past a set force value.

For the mechanical design process, the computer-aided design (CAD) software Solidworks is used to design a novel prosthetic arm. As the main size requirement, measurements were taken from a hand to create the first prototype. Following the prototype design, finite element analysis (FEA), is used for structural analysis to determine the Safety Factor of parts in the assembly due to actuators force, loads and to determine the payload allowance. Simulations show that the prosthetic arm is able to hold a payload force of 45 Newtons (10.1 lbf). This would be the same weight as everyday household objects such as a cup, a bowl, or a book.

Our reinforcement learning grasping algorithm shows promising results in generalizing across multiple everyday objects such as a cuboid, sphere, ellipsoid, cylinder, can, and screwdriver.

We can see similar approaches being applied to other medical assistive devices where a human is interacting with a machine to complete a task. Future work in this area benefits both the users of human-machine interfaces and researchers who are seeking to better understand the links between robot control and advances in machine intelligence.